\documentclass[letterpaper, 10 pt, conference]{ieeeconf}  
\IEEEoverridecommandlockouts                              
\usepackage[colorlinks,linkcolor=green,citecolor=red,urlcolor=blue,bookmarks=true,hypertexnames=true]{hyperref} 

\usepackage{graphicx} 
\usepackage{epsfig} 
\usepackage{mathptmx} 
\usepackage{mathptmx} 
\usepackage{amsmath} 
\usepackage{amssymb}  
\usepackage{booktabs}
\usepackage{url}
\usepackage{multirow}
\usepackage{multicol}

\usepackage{subcaption}
\usepackage[compatibility=false]{caption}
\usepackage{array}                 
\usepackage{float}

\title{\LARGE \bf
End-to-End Generation of City-Scale Vectorized Maps by Crowdsourced Vehicles}

\author{Zebang Feng$^{1}$, Miao Fan$^{2,*}$, Bao Liu$^{1}$, Shengtong Xu$^{3}$, Haoyi Xiong$^{4}$ 
\\
\\
$^{1}$NavInfo Co., Ltd.,
$^{2}$Beijing Institute of Graphic Communication, 
$^{3}$Autohome Inc.,
$^{4}$Baidu Inc.
\thanks{*Corresponding author: Miao Fan (miao.fan@ieee.org), professor at Beijing Institute of Graphic Communication, senior member of IEEE.}
}
        
\begin{document}

\maketitle
\thispagestyle{empty}
\pagestyle{empty}

\begin{abstract}
High-precision vectorized maps are indispensable for autonomous driving, yet traditional LiDAR-based creation is costly and slow, while single-vehicle perception methods lack accuracy and robustness, particularly in adverse conditions. This paper introduces EGC-VMAP, an end-to-end framework that overcomes these limitations by generating accurate, city-scale vectorized maps through the aggregation of data from crowdsourced vehicles. Unlike prior approaches, EGC-VMAP directly fuses multi-vehicle, multi-temporal map elements perceived onboard vehicles using a novel Trip-Aware Transformer architecture within a unified learning process. Combined with hierarchical matching for efficient training and a multi-objective loss, our method significantly enhances map accuracy and structural robustness compared to single-vehicle baselines. Validated on a large-scale, multi-city real-world dataset, EGC-VMAP demonstrates superior performance, enabling a scalable, cost-effective solution for city-wide mapping with a reported 90\% reduction in manual annotation costs.

\end{abstract}
{\keywords vectorized maps, crowdsourced mapping, multi-vehicle perception, end-to-end map generation.}

\begin{figure}[!t]
    \centering
    \includegraphics[width=0.95\columnwidth]{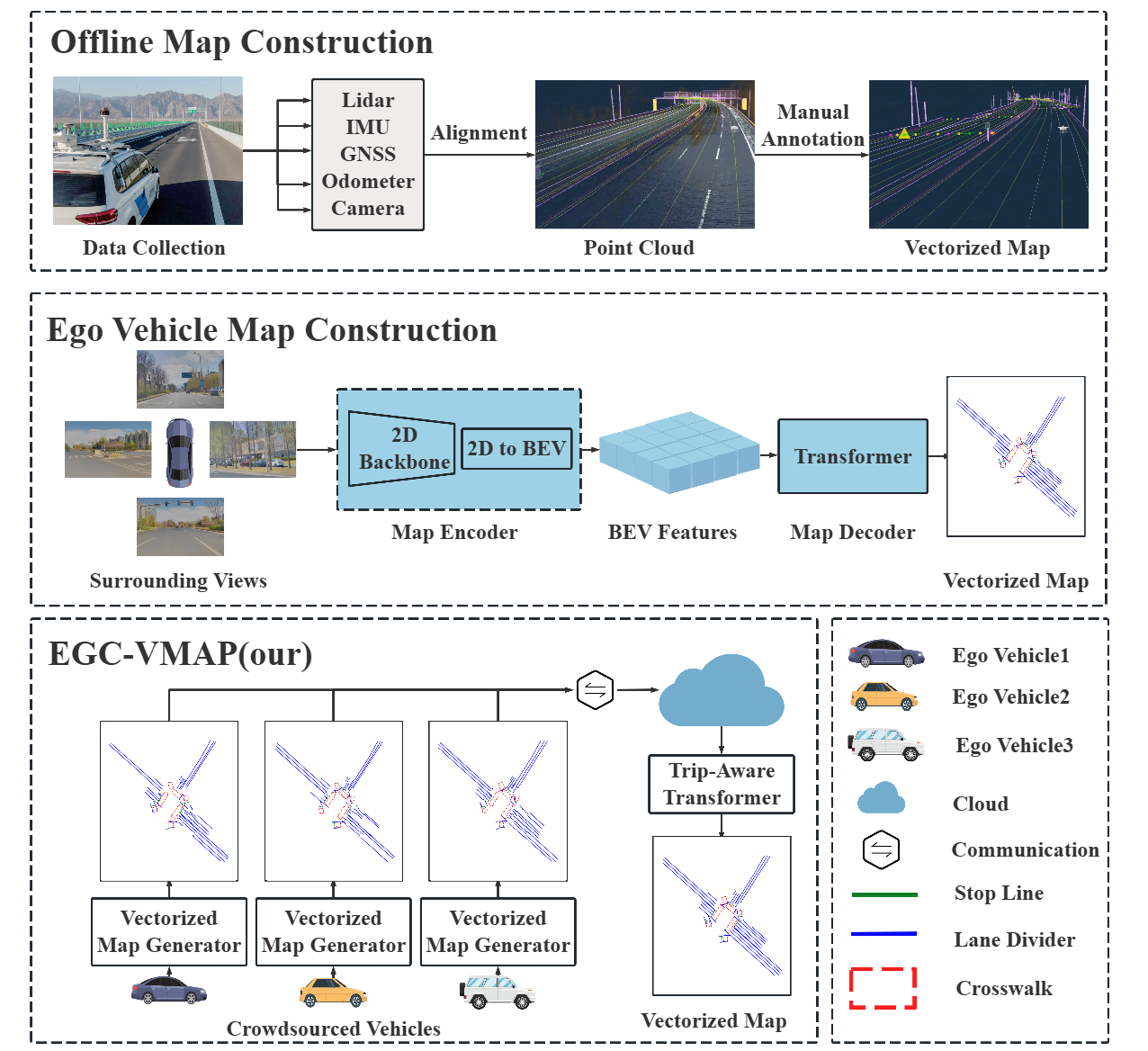}
    \caption{Comparison of three mainstream approaches on vectorized map construction. Offline map construction achieves high accuracy but at a high cost. Ego vehicle map construction offers flexibility but lacks stability. EGC-VMAP proposed by this paper integrates redundant spatiotemporal data, combining cost efficiency with performance advantages.}\vspace{-3mm}
    \label{fig:map_construction}
\end{figure}

\section{Introduction}
Vectorized maps are fundamental to modern autonomous driving systems, providing the essential lane-level detail required for safe navigation, precise localization, and complex behavior planning. However, creating and maintaining these maps at the scale and frequency demanded by dynamic urban environments remains a significant challenge. The dominant paradigm relies on specialized survey vehicles equipped with high-precision LiDAR and other sensors, followed by extensive manual annotation~\cite{survey_slam_hdmap, ziegler2014automatic, apolloscape_hdmap}. While capable of achieving high accuracy, this traditional approach is plagued by prohibitive operational costs, labor-intensive processes, and slow update cycles, rendering it ill-suited for capturing the constant flux of city infrastructure.

To address the cost and scalability limitations, research has explored constructing vectorized maps directly from the perception systems of individual ego-vehicles~\cite{HDMapNet, VectorMapNet, liao2022maptr, liao2024maptrv2, yuan2024streammapnet, wang2024stream, hu2024admap, xiong2024ean, zhou2024himap}. Leveraging onboard sensors like cameras offers a potentially lower-cost, more scalable alternative. However, these methods are inherently constrained by the limited field-of-view of a single vehicle, susceptibility to sensor noise, and significant performance degradation under adverse weather, lighting, or occlusion scenarios. Consequently, maps generated from single ego-vehicles often suffer from geometric inaccuracies, missing elements, and structural inconsistencies, compromising their reliability for safety-critical applications.

Recognizing these approaches' complementary strengths and weaknesses, crowdsourced mapping has emerged as a promising direction~\cite{zhu2023mapprior, jiang2024p, xiong2023neural, he2023vi, hdmapfusion}. By aggregating observations from multiple standard vehicles traversing the same areas over time, it offers the potential to achieve both broad coverage and high fidelity through redundant data fusion. However, effectively integrating this heterogeneous, asynchronous, potentially noisy crowdsourced data into a coherent, accurate map presents unique technical challenges, often involving complex multi-stage processing~\cite{xu2022cobevt, yuan2023generating}.

In this paper, we introduce \textbf{EGC-VMAP} (End-to-End Generation of City-Scale Vectorized Maps by Crowdsourced Vehicles), a novel framework designed to directly tackle these challenges and unlock the potential of crowdsourced mapping (illustrated in Fig.~\ref{fig:map_construction}). Unlike multi-stage pipelines, EGC-VMAP proposes an \emph{end-to-end learning approach} that fuses vectorized map elements, perceived locally onboard individual crowdsourced vehicles, into a unified, high-precision map representation in the cloud. Our core innovation lies in the \emph{Trip-Aware Transformer architecture}~\cite{NIPS2017_3f5ee243}, specifically designed to model the complex spatial and semantic relationships within the aggregated multi-vehicle, multi-temporal data. This is complemented by efficient training mechanisms, including a \emph{hierarchical matching strategy}~\cite{liao2022maptr} to align predictions with ground truth and a \emph{multi-task supervision} scheme that jointly optimizes for classification accuracy, geometric fidelity, and directional consistency. 

While existing works have made significant strides in vectorized map generation, EGC-VMAP introduces several key distinctions from prior work. Unlike single-vehicle end-to-end frameworks (e.g., HDMapNet~\cite{HDMapNet}, VectorMapNet~\cite{VectorMapNet}, MapTR~\cite{liao2022maptr, liao2024maptrv2}, StreamMapNet~\cite{yuan2024streammapnet}) limited by individual perspectives, our approach leverages crowdsourced data redundancy for enhanced reliability and completeness via end-to-end fusion. Furthermore, contrasting with methods that fuse raw sensor data~\cite{hdmapfusion, xu2022cobevt} or rely heavily on priors~\cite{zhu2023mapprior, jiang2024p, xiong2023neural}, EGC-VMAP efficiently operates on \textit{vectorized map elements} generated locally on each vehicle, reducing computational and communication overhead by shifting processing to the edge. This fusion is enabled by our Trip-Aware Transformer~\cite{NIPS2017_3f5ee243}, specifically tailored for aggregating asynchronous, multi-source vector data with explicit trip encoding, differentiating it from generic vision transformers~\cite{DBLP:conf/iclr/DosovitskiyB0WZ21} or standard mapping decoders~\cite{liao2022maptr}. Finally, while components such as hierarchical matching~\cite{liao2022maptr, zhu2020deformable} are adapted, their integration within our unique framework for crowdsourced vector fusion represents a novel application optimized to this specific challenge. A rigorous evaluation of our large-scale released dataset shows that EGC-VMAP significantly outperforms prior work. Successful deployment on the Navinfo platform confirms its practical, cost-effective city-scale mapping capability. The main contributions of this paper are summarized as follows.

\begin{itemize}
    \item \textbf{An End-to-End Crowdsourced Data Collection and Optimization Framework:} We propose EGC-VMAP, the first end-to-end framework, to our knowledge, that directly fuses locally perceived vectorized map elements from multiple crowdsourced vehicles for city-scale high-precision map generation.
    \item \textbf{Trip-Aware Transformer Architecture:} We introduce a specialized Transformer architecture~\cite{NIPS2017_3f5ee243} capable of effectively modeling and integrating geometric and semantic information from asynchronous, multi-source vectorized map data.
    \item \textbf{Effective Learning Strategy:} We employ a combination of hierarchical matching~\cite{liao2022maptr} and multi-task supervision to enable stable and efficient end-to-end training for high-quality map element prediction.
    \item \textbf{Industry Practices and Open-source Contribution:} Extensive experiments on a new large-scale dataset and successful deployment validate the significant performance gains and cost-reduction potential of our approach for real-world, city-scale mapping. Code and our massive data collection are open-sourced\footnote{\url{https://github.com/flame4343/EGC-VMAP}}.
\end{itemize}




\section{Related Work}
This section briefly reviews vectorized maps generation from two perspectives: (1) map construction methods and (2) crowdsourced mapping. 
\begin{itemize}
    \item {\bf Map Construction Methods:} Early methods relied on costly survey vehicles with multi-sensor systems (LiDAR, IMU, GNSS) and extensive manual annotation or rule-based extraction~\cite{survey_slam_hdmap, ziegler2014automatic, apolloscape_hdmap}, achieving high accuracy but suffering from slow updates and limited scalability. To improve efficiency, end-to-end approaches emerged, leveraging onboard perception, often in the Bird's-Eye View (BEV) space. Preliminaries such as HDMapNet~\cite{HDMapNet} used parallel branches and post-processing, while VectorMapNet~\cite{VectorMapNet} introduced autoregressive decoding. MapTR~\cite{liao2022maptr, liao2024maptrv2} significantly advanced the field with hierarchical matching, inspiring subsequent studies~\cite{hu2024admap, xiong2024ean, zhou2024himap}. Temporal consistency was further explored by methods such as StreamMapNet~\cite{yuan2024streammapnet} and SQD~\cite{wang2024stream}. Despite automation, these single-vehicle methods remain vulnerable to limited fields-of-view and occlusions. 
    \item {\bf Crowdsourced Mapping Methods:} Crowdsourced mapping offers a promising alternative by aggregating multi-source, spatiotemporally diverse data to enhance stability and update frequency. Some approaches integrate prior map information~\cite{zhu2023mapprior, jiang2024p, xiong2023neural, ho2024mapex} or leverage infrastructure data~\cite{he2023vi, yang2023vilam}. Others focus on fusing heterogeneous sensor data from multiple vehicles~\cite{xu2022cobevt, yuan2023generating, mehr2019disconet} to compensate for single-vehicle limitations. However, many crowdsourced methods still rely on multi-stage pipelines or face challenges with heterogeneous data distributions and complex matching, hindering unified end-to-end optimization.
\end{itemize}

\begin{figure*}[!h]
    \centering
    \includegraphics[width=0.95\textwidth]{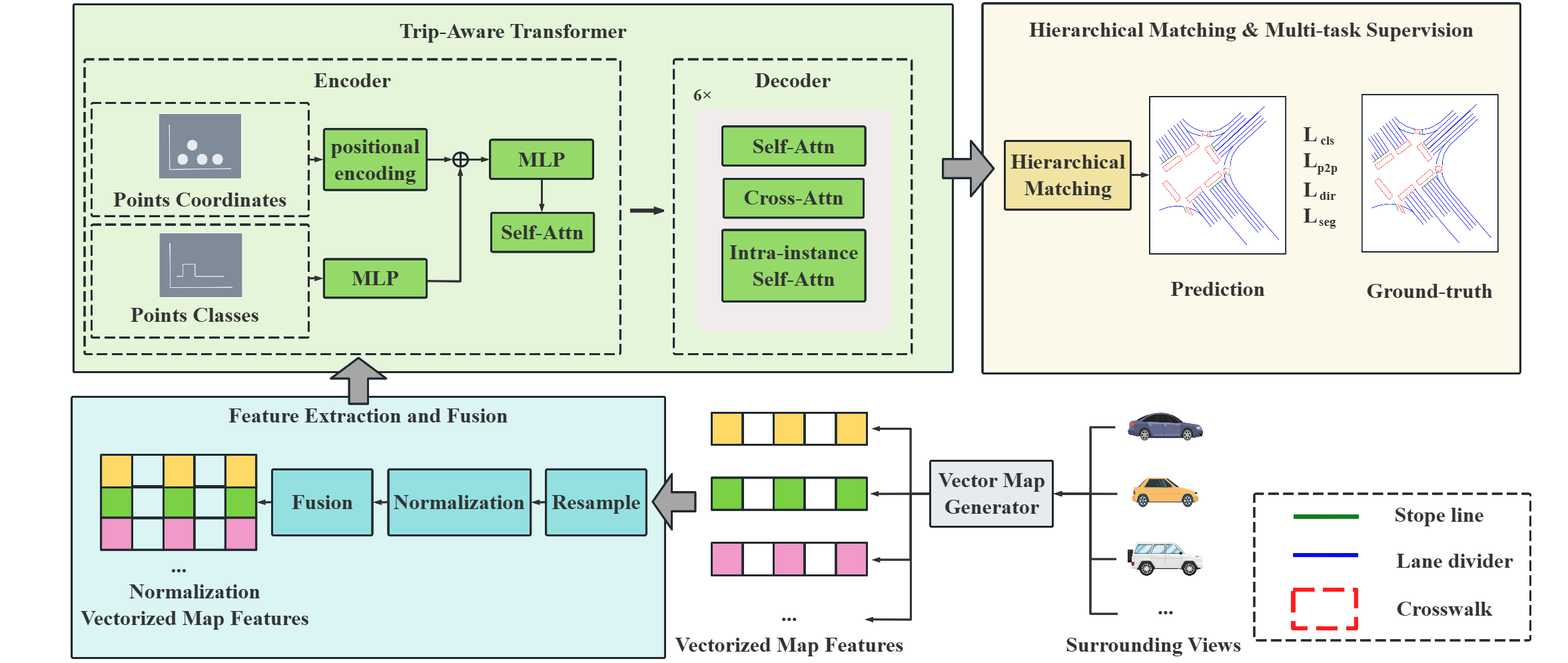}
    \caption{Overview of the EGC-VMAP framework.
    It aggregates crowdsourced vectorized map data into a unified input representation. 
    A trip-aware Transformer encodes features and performs conditional query decoding to generate normalized vectorized representations of map elements. 
    Hierarchical matching and multi-task supervision are used jointly to enable the efficient and accurate end-to-end construction of city-scale vector maps.}\vspace{-3mm}
    \label{fig: pipeline}
\end{figure*}


\section{Problem Formulation}
This section formally defines the task of generating city-scale vectorized maps from crowdsourced data.

\subsection{Data Representation}
We adopt vectorized representations for map data:
\begin{itemize}
    \item \textbf{Map Element ($V$):} A single semantic element on the map (e.g., a lane divider, stop line). It is represented as a set of $N_p$ ordered points with associated class labels:  $V = \{ (P_n, c_n) \}_{n=0}^{N_p-1}$, where $P_n \in \mathbb{R}^2$ denotes the geographical coordinates (e.g., $(x, y)$) of the $n$-th point, and $c_n$ represents the class label of the element (often constant for all points within an element, $c_n=c$). $N_p$ is the number of points defining the element.

    \item \textbf{Single-Vehicle Perceived Map ($E$):} The vectorized map generated by a single crowdsourced vehicle during one trip or observation period. It consists of $N_v$ map elements: $E = \{V_m^{\text{sv}}\}_{m=0}^{N_v-1}$, where $N_v$ is the number of map elements perceived by that vehicle.

    \item \textbf{Crowdsourced Map Data ($X$):} The collection of multiple, asynchronous perceived maps obtained from $N_e$ different crowdsourced vehicle trips covering the same geographical area: $X = \{E_k\}_{k=0}^{N_e-1}$, where $N_e$ is the number of individual perceived maps in the collection.

    \item \textbf{Ground Truth Map ($Y$):} The high-precision reference vectorized map for the target area, typically constructed using traditional survey methods. It shares the same data format as a single-vehicle map $E$: $Y = \{V_i\}_{i=0}^{N_{gt}-1}$, where $N_{gt}$ is the number of ground truth map elements.
\end{itemize}

\subsection{Learning Objective}
The overall goal is to learn a mapping function $F_\theta$, parameterized by learnable parameters $\theta$, that takes the collection of potentially noisy and incomplete crowdsourced maps $X$ as input and outputs an accurate, fused vectorized map $\hat{Y}$ that closely approximates the ground truth map $Y$:
\begin{equation}
    \hat{Y} = F_\theta(X)
\end{equation}
The objective is to minimize the discrepancy between the predicted map $\hat{Y}$ and the ground truth map $Y$.




\section{Method}
This section provides a detailed description of the proposed EGC-VMAP framework. The framework enhances the accuracy of vectorized map prediction by integrating vectorized maps generated from crowdsourced vehicle perceptions. The overall architecture of the EGC-VMAP framework is illustrated in Fig.~\ref{fig: pipeline} and comprises three main components:
(1) Feature Extraction and Fusion;
(2) Trip-Aware Transformer;
(3) Hierarchical Matching and Multi-task Supervision.
The following subsections offer an in-depth explanation of each component.


\subsection{Feature Extraction}

To fully exploit and integrate crowdsourced vehicle perception data, we design a unified feature extraction and fusion mechanism that generates feature representations with rich structural information and high stability, thereby providing strong support for subsequent vectorized map construction tasks.

Specifically, the input consists of $N_e$ crowdsourced vectorized maps, where each map contains $N_v$ map elements, and each element is composed of $N_p$ geographic coordinate points. Each coordinate point records its spatial location $(x, y)$ and the corresponding semantic category label $c$. Based on this, the initial spatial coordinate features are defined as $\mathbf{X}_{\text{coord}} \in \mathbb{R}^{N_e \times N_v \times N_p \times 2}$, and the category label features are defined as $\mathbf{X}_{\text{cls}} \in \mathbb{R}^{N_e \times N_v \times N_p \times 1}$.

Considering the significant variations in sampling point density and length among map elements, we apply an arc-length-based uniform resampling method to preprocess the original map elements. Meanwhile, we perform min-max normalization on all spatial coordinates to mitigate the influence of differing map scales. This module provides high-quality feature inputs for the subsequent Trip-Aware Transformer encoder, further enhancing the model’s overall performance and generalization capability.







\subsection{Trip-Aware Transformer}
\label{sec:trip-aware-transformer}
\subsubsection{Encoder}
We adopt a simple yet effective architecture to fuse geometric and semantic information across different vectorized map features. Specifically, a multi-layer perceptron (MLP) network is employed to encode the semantic information of each map element, while positional encoding is used to encode the geometric information. In positional encoding, we explicitly incorporate embeddings of both the vectorized map and the positions of individual map elements. After concatenating the semantic and geometric encodings, the features are mapped to a high-dimensional feature space through another MLP. Finally, a self-attention mechanism is applied to model global features, capturing spatial and semantic correlations across different trips and temporal observations. This process provides rich contextual information to support the subsequent decoding phase.


\subsubsection{Decoder} 
We design a novel query composition to flexibly encode structured map information and perform hierarchical bipartite matching for learning map elements. Specifically, we extend the hierarchical query embedding scheme introduced in MapTR~\cite{liao2022maptr} and customize two types of queries. In the map element queries, multiple cascaded decoding layers are used to produce instance-level queries and point-level queries that are shared across all instances. Inspired by the idea from DuMapNet~\cite{DuMapNet}, we further introduce foreground queries to support an auxiliary semantic segmentation task, which improves prediction accuracy and accelerates training convergence.

The decoder is composed of multiple cascaded layers, each containing a self-attention module, a cross-attention module, and an intra-instance self-attention module. The self-attention module enables hierarchical queries to exchange information across the entire feature space. The cross-attention module facilitates interactions between hierarchical queries and spatial features. The intra-instance self-attention module enhances interactions among points within the same instance.

The output of the query embeddings encapsulates the point set representation of each individual map element. We adopt shared classification and point regression heads to process the point sets. The classification head predicts the instance category scores, while the point regression head predicts the positions of the points. The model outputs a $\mathbb{R}^{N_p\times2}$ vector for each map instance, representing the normalized coordinates of $N_p$ points. Additionally, we propose a foreground segmentation branch that works in conjunction with the hierarchical queries. After being processed by a conventional decoder network and an MLP encoder, it generates a semantic segmentation map.




\subsection{Hierarchical Matching and Multi-task Supervision}

We define $Y = \{V_i\}_{i=0}^{N_{gt}-1}$ as the ground-truth vectorized map, where $N_{gt}$ denotes the number of map elements. For each map element, we represent it as $V_i = (P_i, c_i)$, where $P_i$ and $c_i$ denote the ground-truth point set and class label of map element $V_i$, respectively. Meanwhile, we define the predicted map elements as $\hat{V}_i = (\hat{P}_i, \hat{c}_i)$, where $\hat{P}_i$ and $\hat{c}_i$ correspond to the predicted point set and classification score, respectively.


\subsubsection{Hierarchical matching}

Following the hierarchical matching strategy proposed in MapTR~\cite{liao2022maptr}, we perform detailed matching at both the instance and point levels. First, we conduct Instance-Level Matching. Specifically, for a given ground-truth map element $V_i$, we seek the corresponding predicted map element $\hat{V}_{\pi(i)}$. Here, $\hat{\pi}$ represents a permutation of the $N_{gt}$ predicted map elements that minimizes the overall instance matching cost, formulated as:
\begin{equation}
\hat{\pi} = \arg\min_{\pi \in \Pi_N} \sum_{i=0}^{N_{gt}-1} L_{\text{match}}^{\text{ins}}(\hat{V}_{\pi(i)}, V_i)
\end{equation}
where the instance-level matching cost is defined by:
\small
\begin{equation}
L_{\text{match}}^{\text{ins}}(\hat{V}_{\pi(i)}, V_i) = L_{\text{Focal}}(\hat{c}_{\pi(i)}, c_i) + L_{\text{position}}(\hat{P}_{\pi(i)}, P_i)
\end{equation}
\normalsize
Here, $L_{\text{Focal}}()$ and $L_{\text{position}}()$ represent the classification matching loss and the position matching loss, respectively. The optimal assignment is obtained using the Hungarian algorithm, following the approach in DETR~\cite{zhu2020deformable}.

Subsequently, we perform Point-Level Matching. For each matched map element $\hat{V}_{\pi(i)}$, we search for the optimal point permutation $\hat{\gamma}$ from the set of all permutations $\Gamma$ that minimizes the overall point matching cost:
\begin{equation}
\hat{\gamma} = \arg\min_{\gamma \in \Gamma} \sum_{j=0}^{N_p-1} D_{\text{Mht}}(\hat{p}_j, p_{\gamma(j)})
\end{equation}
where $D_{\text{Mht}}$ denotes the Manhattan distance.



\subsubsection{Multi-task supervision}

Based on the matching results, the overall loss function consists of four components: classification loss, point-to-point loss, edge direction loss, and foreground segmentation loss, formulated as:
\begin{equation}
L = \alpha_c L_{\text{cls}} + \alpha_p L_{\text{p2p}} + \alpha_d L_{\text{dir}} + \alpha_s L_{\text{seg}}
\end{equation}
where $\alpha_c$, $\alpha_p$, $\alpha_d$, and $\alpha_s$ are hyperparameters controlling the relative weights of each loss term; $L_{\text{cls}}$ denotes the classification loss, $L_{\text{p2p}}$ the point-to-point positional loss, $L_{\text{dir}}$ the edge direction loss, and $L_{\text{seg}}$ the foreground segmentation loss, which is detailed in Section~\ref{sec:trip-aware-transformer}.

Based on the previously defined Focal Loss, the classification loss $L_{\text{cls}}$ is formulated as:
\begin{equation}
L_{\text{cls}} = \sum_{i=0}^{N_{gt}-1} L_{\text{Focal}}(\hat{c}_{\hat{\pi}(i)}, c_i)
\end{equation}

For the ground-truth instance $V_i$, we identify its best-matched predicted instance $\hat{P}_{\hat{\pi}(i)}$. For a point indexed by $j$ in $\hat{P}_{\hat{\pi}(i)}$, we find its optimal corresponding point $\hat{\gamma}_i(j)$ from the ground-truth point set. The point-to-point loss $L_{\text{p2p}}$ is defined as:
\begin{equation}
L_{\text{p2p}} = \sum_{i=0}^{N_{gt}-1} \sum_{j=0}^{N_p-1} D_{\text{Mht}}(\hat{P}_{\hat{\pi}(i),j}, P_{i,\hat{\gamma}_i(j)})
\end{equation}

Since the point-to-point loss only focuses on the nodes of polylines and polygons, we introduce an edge direction loss to further enhance the fitting of edges. Here, the predicted edge is denoted as $\hat{e}_{\hat{P}_{\hat{\pi}(i)},j}$, and the ground-truth edge is denoted as $e_{P_i,\hat{\gamma}_i(j)}$. The edge direction loss is formulated as:
\begin{equation}
L_{\text{dir}} = -\sum_{i=0}^{N_{gt}-1} \sum_{j=0}^{N_p-1} \text{CosSim} \left( \hat{e}_{\hat{P}_{\hat{\pi}(i)},j}, e_{P_i,\hat{\gamma}_i(j)} \right)
\end{equation}

By integrating these three modules, EGC-VMAP achieves efficient, accurate, and robust construction of city-scale vectorized maps. Compared to conventional methods, our approach better adapts to dynamically changing urban environments while maintaining low costs and offering higher mapping accuracy.






\section{Experiment}
To validate the effectiveness of our proposed method for constructing city-scale crowdsourced maps, we conduct a comprehensive evaluation using a large-scale, self-collected real-world map dataset. Our evaluation includes performance comparisons, ablation studies, and visualization analysis.


\subsection{Experimental Setup}
\subsubsection{Real-world dataset}  
To verify the effectiveness of our proposed method in real-world scenarios, we collect and publicly release a large-scale city-level crowdsourced vectorized map dataset, named the Navinfo Dataset. This dataset covers multiple representative urban areas and contains diverse vectorized map data collected by crowdsourced vehicles, along with high-precision ground-truth annotations. In particular, all ground-truth labels in the dataset are collected by Navinfo using specialized vehicles equipped with LiDAR sensors, ensuring high accuracy and reliability of the annotations.  
Regarding the scale of the Navinfo Dataset, data were collected across 10,000 typical scenes in six major cities, covering a variety of road types, including urban arterials, intersections, and service roads. The selected cities—Beijing, Shanghai, Guangzhou, Shenzhen, Chongqing, and Tangshan—were chosen for their diverse city scales and geographical characteristics. In each scene, vectorized map data from at least 10 different crowdsourced vehicles, collected at different time points, is included. The data records contain the vehicle ID, geographic positions based on the WGS84 coordinate system, and various types of map elements, such as lane dividers, stop lines, and crosswalks.

\subsubsection{Evaluation metrics}  
We evaluate the proposed method’s mapping performance on the Navinfo Dataset, following the evaluation framework established by StreamMapNet~\cite{yuan2024streammapnet} on the NuScenes dataset. We focus on three core types of map elements: Lane Dividers, Stop Lines, and Crosswalks. The spatial mapping range is defined as: $X \in [-30.0\,\text{m}, +30.0\,\text{m}]$, $Y \in [-30.0\,\text{m}, +30.0\,\text{m}]$. We adopt Average Precision (AP) and mean Average Precision (mAP) as the primary evaluation metrics to quantify model performance on vectorized map element detection tasks. We introduce the Chamfer Distance $D_{\text{Chamfer}}$ to measure the alignment between predictions and ground-truth labels. Specifically, we compute $AP_\tau$ at multiple thresholds $\tau \in T$ and define the final AP as the average overall thresholds:
\begin{equation}
AP = \frac{1}{|T|} \sum_{\tau \in T} AP_\tau
\end{equation}
where the threshold set is defined as $T = \{0.5, 1.0, 1.5\}$ (in meters), representing the tolerance levels for Chamfer Distance matching.

The AP values for the three types of map elements are denoted as $AP_{\text{Lane}}$, $AP_{\text{Stop}}$, and $AP_{\text{Cross}}$, respectively. The final mAP is defined as:
\begin{equation}
\text{mAP} = \frac{AP_{\text{Lane}} + AP_{\text{Stop}} + AP_{\text{Cross}}}{3}
\end{equation}

\subsubsection{Implementation details}  
Our method is trained on 4 NVIDIA RTX 3090 GPUs with a batch size of 8. We use the AdamW optimizer with an initial learning rate of $1\times10^{-4}$, a weight decay of 0.01, and apply cosine learning rate decay.  
Regarding the model architecture, in the Trip-Aware Transformer module, the number of instance queries, point queries, and decoder layers is set to 30, 30, and 6, respectively.

\subsection{Evaluation} 

\subsubsection{Base models}  
To comprehensively evaluate the effectiveness and generalization capability of our proposed method, we adopt StreamMapNet as the primary baseline model for comparison. StreamMapNet has demonstrated outstanding performance in vectorized map generation tasks, making it a strong baseline candidate. Furthermore, to further validate the applicability and performance gains of our approach, we integrate the proposed EGC-VMAP into four representative camera-based vectorized map generation methods: HDMapNet, VectorMapNet, MapTR, and StreamMapNet, as summarized in Table~\ref{tab: comparison}. We maintain the original designs and hyperparameter settings of each baseline model to ensure fair comparisons. By incorporating EGC-VMAP into these mainstream frameworks, we systematically evaluate its compatibility and performance improvements across different vectorized map generation pipelines.

\begin{table}[!t]
    \centering
    \scriptsize
    \caption{Quantitative analysis of vectorized map generation. EGC-VMAP improves these methods by aggregating multi-source vehicle perception data.}
    \label{tab: comparison}
    \begin{tabular}{l|rrrrr}
    \toprule
    \multirow{2}{*}{\textbf{Model}} & \multicolumn{4}{c}{\textbf{Average Precision}} \\ 
                          & $AP_{\text{Lane}}$  & $AP_{\text{Stop}}$  & $AP_{\text{Cross}}$ & mAP \\ 
    \midrule
    \textbf{HDMapNet}                              &20.62 &21.08 &13.13 &18.28 \\ 
    \textbf{HDMapNet + EGC-VMAP}                   &25.33 &25.99 &18.27 &23.20  \\ 
    \textbf{VectorMapNet}                          &25.49 &26.67 &18.85 &23.00 \\ 
    \textbf{VectorMapNet + EGC-VMAP}               &30.88 &31.72 &23.30 &28.63 \\ 
    \textbf{MapTR}                                 &41.51 &43.37 &18.16 &34.35 \\ 
    \textbf{MapTR + EGC-VMAP}                      &45.67 &47.52 &22.11 &38.43 \\ 
    \textbf{StreamMapNet}                          &48.12 &49.67 &24.33 &40.71 \\ 
    \textbf{StreamMapNet + EGC-VMAP}               &52.88 &54.34 &29.34 &45.52 \\ 
    \bottomrule
    \end{tabular}
\end{table}

\begin{table}[!t]
    \centering
    \caption{Crowdsourced vehicle data fusion.}
    \label{table: Data Fusion}
    \begin{tabular}{l|rrrrr}
    \toprule
    \multirow{2}{*}      {\textbf{Data Fusion}} & \multicolumn{4}{c}{\textbf{Average Precision}} \\ 
                          &$AP_{\text{Lane}}$  &$AP_{\text{Stop}}$ &$AP_{\text{Stop}}$ &mAP \\ 
    \midrule
    \textbf{Baseline}                              &48.12 &49.67 &24.33 &40.71  \\  
    \textbf{Ego Vehicle}                           &49.01 &50.13 &25.41 &41.52  \\ 
    \textbf{Crowdsourced Vehicle}                 &52.88 &54.34 &29.34 &45.52  \\ 
    \bottomrule
    \end{tabular}
\end{table}

\begin{table}[!t]
    \centering
    \caption{EGC-VMAP performance under adverse environments.}
    \label{table: Adverse Environments}
    \begin{tabular}{l|rrrrr}
    \toprule
    \multirow{2}{*}      {\textbf{Environments}} & \multicolumn{4}{c}{\textbf{Average Precision}} \\ 
                          &$AP_{\text{Lane}}$  &$AP_{\text{Stop}}$ &$AP_{\text{Stop}}$ &mAP \\ 
    \midrule
    \textbf{Normal}               &50.30 &51.22 &26.14 &42.55    \\ 
    \textbf{Normal + EGC-VMAP}    &54.12 &56.86 &31.16 &47.38    \\ 
    \textbf{Rain}                 &46.54 &47.52 &22.04 &38.70    \\  
    \textbf{Rain + EGC-VMAP}      &50.72 &52.96 &27.17 &43.62    \\ 
    \textbf{Night}                &43.67 &44.42 &19.09 &35.73    \\  
    \textbf{Night + EGC-VMAP}     &48.79 &50.91 &25.37 &41.69    \\ 

    \bottomrule
    \end{tabular}
\end{table}
\begin{table*}[t!]
    \centering
    \caption{Ablation study on the key components. }
    \label{tab: ablation}
    \begin{tabular}{l|ccccrrrr}
    \toprule
    \multirow{2}{*}{\textbf{Model}} & \multicolumn{4}{c}{\textbf{Component}} & \multicolumn{4}{c}{\textbf{Average Precision}} \\
    & \textit{Small-scale Fusion} & \textit{Large-scale Fusion} & \textit{Positional Encoding}  & \textit{Segmentation Branch} &$AP_{\text{Lane}}$  &$AP_{\text{Stop}}$ &$AP_{\text{Stop}}$ & \textit{mAP} \\
    \midrule
    A &\checkmark &           &           &           &49.12 &50.67 &25.33 &41.71  \\
    B &           &\checkmark &           &           &50.88 &51.51 &25.68 &42.69  \\
    C &           &\checkmark &\checkmark &           &51.53 &52.37 &27.20 &43.70  \\
    D &           &\checkmark &           &\checkmark &52.06 &53.11 &28.99 &44.72  \\
    E &           &\checkmark &\checkmark &\checkmark &52.88 &54.34 &29.34 &45.52  \\
    \bottomrule
    \end{tabular}
\end{table*}
\begin{figure*}[ht!]
    \centering
    \includegraphics[width=0.78\textwidth]{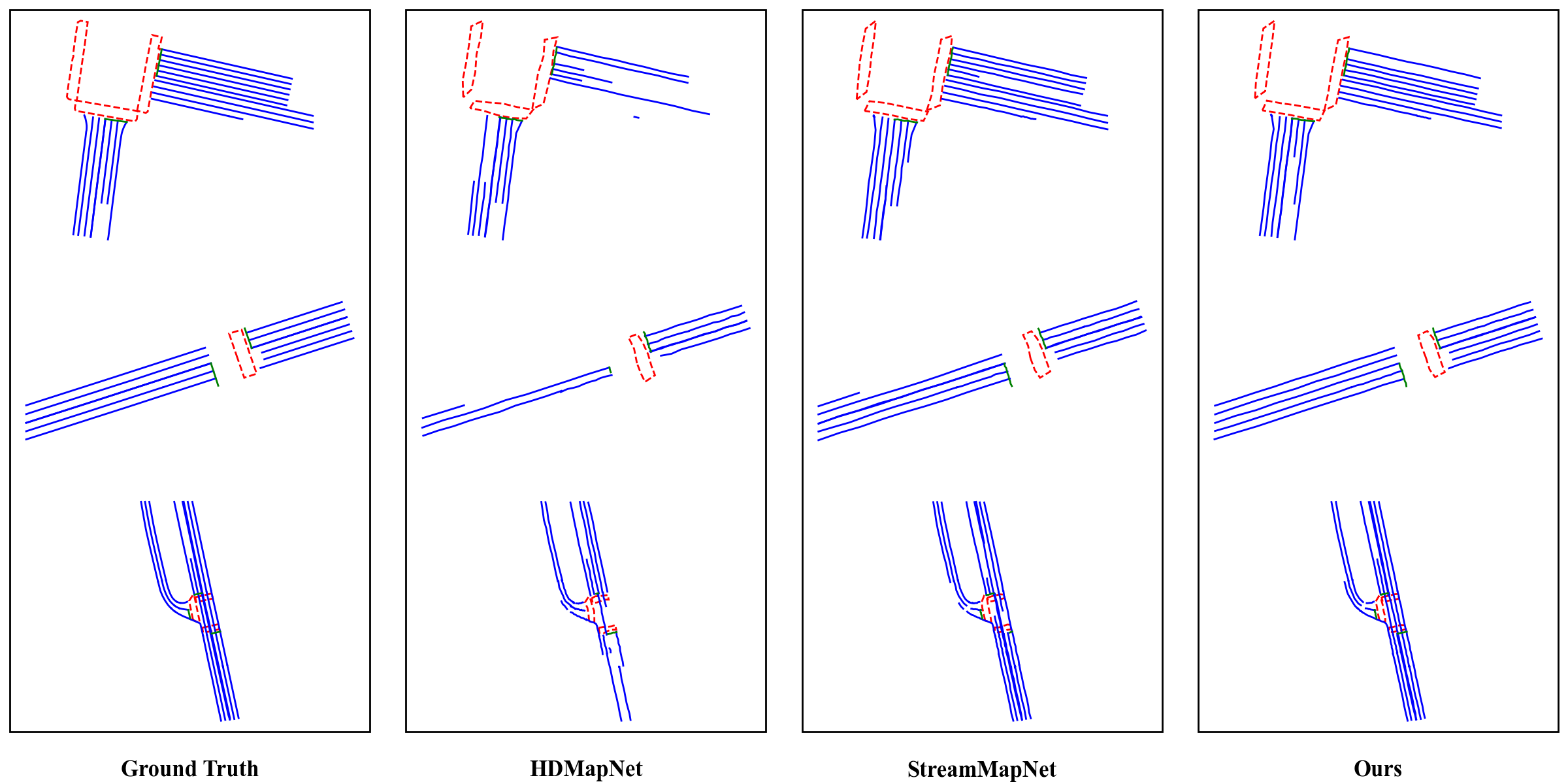}
    \caption{Qualitative results.
    Comparison of our method with several existing methods on the Navinfo dataset.
    From the first to the fourth column: Ground truth, HDMapNet, StreamMapNet, StreamMapNet with EGC-VMAP.
    In these results, lane dividers, stop lines, and crosswalks are represented in blue, green, and red, respectively.}
    \label{fig: vis}
    \vspace{-3mm}
\end{figure*}
\subsubsection{Data fusion by crowdsourced vehicles}  
As shown in Table~\ref{table: Data Fusion}, we present the improvements in map construction accuracy achieved by fusing multiple vectorized maps collected from different crowdsourced vehicles.  
In our experiments, “Ego Vehicle” denotes fusion using maps generated by a single vehicle during a trip, whereas “Crowdsourced Vehicle” refers to the fusion of maps collected from different vehicles observing the same scene.  
The results demonstrate that aggregating data from crowdsourced vehicles improves mapping accuracy.  
Compared to single-vehicle fusion, crowdsourced fusion spans a broader temporal range and exhibits greater reliability to outlier data, leading to a notable mAP improvement of 4 points. This highlights the effectiveness of leveraging multi-vehicle observations for constructing more accurate and reliable maps.
\vspace{-1mm}
\subsubsection{Performance under adverse environments}  
As shown in Table~\ref{table: Adverse Environments}, EGC-VMAP consistently outperforms baseline methods under various conditions. 
While single-vehicle perception degrades in adverse scenarios such as rain and nighttime due to reduced image quality, EGC-VMAP mitigates these issues by fusing multi-vehicle, multi-temporal data. 
It improves mAP by 4.83, 4.92, and 5.96 points under normal, rain, and nighttime scenarios, demonstrating strong robustness and generalization capability.




\subsection{Ablation Study}
To assess the contribution of each key module within our model, we design the following ablation experiments:
\begin{itemize}
    \item \textbf{Small-scale Fusion}: Fusing at most three samples to evaluate the model’s performance under limited data availability.
    \item \textbf{Large-scale Fusion}: Fusing ten samples to investigate the performance gains from incorporating a larger amount of data.
    \item \textbf{Positional Encoding}: Introducing trip-aware positional embeddings to enhance the model’s spatial understanding of specific locations.
    \item \textbf{Segmentation Branch}: Adding an auxiliary semantic segmentation task to improve prediction accuracy and accelerate training convergence.
\end{itemize}

Table~\ref{tab: ablation} summarizes the performance comparisons under different settings.  
Experimental results demonstrate that the model’s performance improves significantly with increased fused samples. 
Specifically, moving from Small-scale Fusion (A) to Large-scale Fusion (B) results in a mAP improvement of 0.98 points.
Adding the Positional Encoding Module (C) further enhances the mAP by 1.01 points, highlighting its crucial role in strengthening spatial perception, particularly in complex scenes.
Meanwhile, introducing the Segmentation Branch (D) boosts the mAP by 2.03 points, enabling more fine-grained, pixel-level semantic modeling and substantially improving semantic understanding and prediction quality.
Finally, incorporating the Positional Encoding and Segmentation Branch (E) performs best.
These results verify each component's effectiveness in progressively enhancing spatial reasoning and semantic prediction capabilities.



\subsection{Visualization Analysis}
Fig.~\ref{fig: vis} presents the mapping results produced by our method across various urban environments. Our approach successfully generates structurally continuous and semantically coherent vectorized lane lines as illustrated. Moreover, our vectorized lane lines exhibit significantly improved recall and precision compared to other methods. 
Our method also maintains excellent performance in challenging scenarios such as sharp turns, intersections, and areas with occlusions, demonstrating its reliability and strong generalization capability.


\section{Conclusion}

This paper introduced EGC-VMAP, an end-to-end framework addressing cost, scalability, and robustness challenges in city-scale vectorized mapping by directly fusing locally perceived vector elements from crowdsourced vehicles. Our core technical contributions include the \emph{Trip-Aware Transformer architecture} for modeling asynchronous multi-vehicle data, combined with \emph{hierarchical matching} and a \emph{multi-task loss} optimizing classification ($L_{cls}$), point accuracy ($L_{p2p}$), edge direction ($L_{dir}$), and segmentation ($L_{seg}$), all within an \emph{end-to-end crowdsourced data collection and optimization framework}. Extensive experiments on our large-scale dataset validate the efficacy of EGC-VMAP. It significantly enhances map accuracy, achieving a notable \emph{4.81 mAP} improvement over the StreamMapNet baseline and demonstrating a \emph{4 mAP} gain from multi-vehicle crowdsourcing versus single-vehicle data. Furthermore, EGC-VMAP exhibits enhanced robustness, yielding substantial gains in adverse conditions (\emph{+4.92 mAP} in rain, \emph{+5.96 mAP} at night). These results confirm the capability of  EGC-VMAP to produce accurate, reliable maps cost-effectively, offering a practical path towards scalable, frequently updated city-scale maps. Future work will focus on enhancing real-time performance, improving accuracy on complex topologies, and expanding support for a broader range of road element types to handle diverse and dynamic urban environments better.



\bibliographystyle{IEEEtran}
\bibliography{IEEEexample}

\end{document}